\documentclass[numbered,3p,sort&compress]{elsarticle} 

\usepackage{natbib}
\usepackage{geometry}
\usepackage{dsfont}
\usepackage{amsmath}
\usepackage{cancel}	
\usepackage{amssymb}
\usepackage{amsbsy}
\usepackage{amsfonts}
\usepackage{latexsym}
\usepackage{bm}
\usepackage{verbatim}
\usepackage{mathrsfs}
\usepackage{graphicx}
\usepackage{color}
\usepackage{subfigure}
\usepackage{wrapfig}
\usepackage{float}

\usepackage[labelfont=bf,justification=justified,singlelinecheck=true,font=normal]{caption} 

\journal{the IEEE Transactions on Ultrasonics, Ferroelectrics and Frequency Control}

\definecolor{red}{rgb}{1,0,0}								
\definecolor{blue}{rgb}{0,0,1}							

\newcommand{\jdh}[1]{}
\renewcommand{\jdh}[1]{{\color{black}{#1}}} 

\begin{document}
\title{Automated Defect Localization via Low Rank Plus Outlier Modeling of Propagating Wavefield Data} 


\author[MINNce]{Stefano Gonella~\corref{cor1}} 
\ead{sgonella@umn.edu}

\author[MINNece]{Jarvis D. Haupt}
\ead{jdhaupt@umn.edu}

\address[MINNce]{Department of Civil Engineering, University of Minnesota, Minneapolis, MN}
\address[MINNece]{Department of Electrical and Computer Engineering, University of Minnesota, Minneapolis, MN}

\cortext[cor1]{\textbf{Corresponding Author}: 500 Pillsbury Drive S.E. Minneapolis, MN 55455-0116. Phone (612) 625-0866}



\begin{abstract}
This work proposes an agnostic inference strategy for material diagnostics, conceived within the context of laser-based non-destructive evaluation methods which extract information about structural anomalies from the analysis of acoustic wavefields measured on the structure's surface by means of a scanning laser interferometer. The proposed approach couples spatiotemporal windowing with low rank \jdh{plus outlier modeling}, to identify a priori unknown deviations in the propagating wavefields caused by material inhomogeneities or defects, using virtually no knowledge of the structural and material properties of the medium.  This characteristic makes the approach particularly suitable for diagnostics scenarios where the mechanical and material models are complex, unknown, or unreliable. We demonstrate our approach in a simulated environment using benchmark point and line defect localization problems based on propagating flexural waves in a thin plate.
\end{abstract}


\maketitle

\section{Introduction} \label{Introduction}

Within the realm of dynamics-based non-destructive evaluation (NDE) and damage prognosis methodologies, techniques based on guided waves have become popular~\cite{Staszewski,Rose, Giurgiutiu_book} due to their sensitivity to a variety of damage types and their ability to travel long distances (with minor attenuation) and interact with defects located far from available actuation and sensing points. Guided waves are generated and received by transmitter-receiver pairs distributed over the test specimen and the detection process follows the pitch-catch ~\cite{Pitch_Catch_Fu_Kuo_Chang} or the pulse-echo~\cite{Raghavan_Cesnick_Echo_pulse_2008} paradigms: a signature of wave scattering is captured along the transmitter-receiver path, and the position of the defect is triangulated from the data collected from multiple transducer pairs.
Numerous efforts have been devoted to the construction of \jdh{techniques for damage estimation, classification and localization}: some popular approaches involve statistical methods~\cite{Likelihood_Guided_Waves_Flynn}, acoustic imaging techniques~\cite{Michaels_Sparse}, singular value decomposition~\cite{SVD_Sensors_Damage_NCState}, neural networks~\cite{Lamb_waves_delamination_Neural}, conventional and Monte Carlo matching pursuit decomposition~\cite{Das_et_al_MPD_SPIE,Das_Chattopadhyay_MoteCarlo_MPD,Mallat_MPD_Dictionaries} and support vector machines~\cite{Das_et_al_SPIE_SVM}, among others. \jdh{Damage detection and triangulation has also been successfully tackled using delay-and-sum techniques \cite{Michaels07WaMot, Michaels08SMS} and phased arrays~\cite{Yu_Giurgiutiu_Jomms_2007}}. Spatial optimization of the sensor networks has also been proposed~\cite{Wang_Yuan_PZT_Locations_Damage_Detection} to enhance the detection capabilities.

\begin{figure*}[t]
\centering
		\subfigure[Material damping]{\label{Effect_of_damping} \includegraphics[scale=0.6]{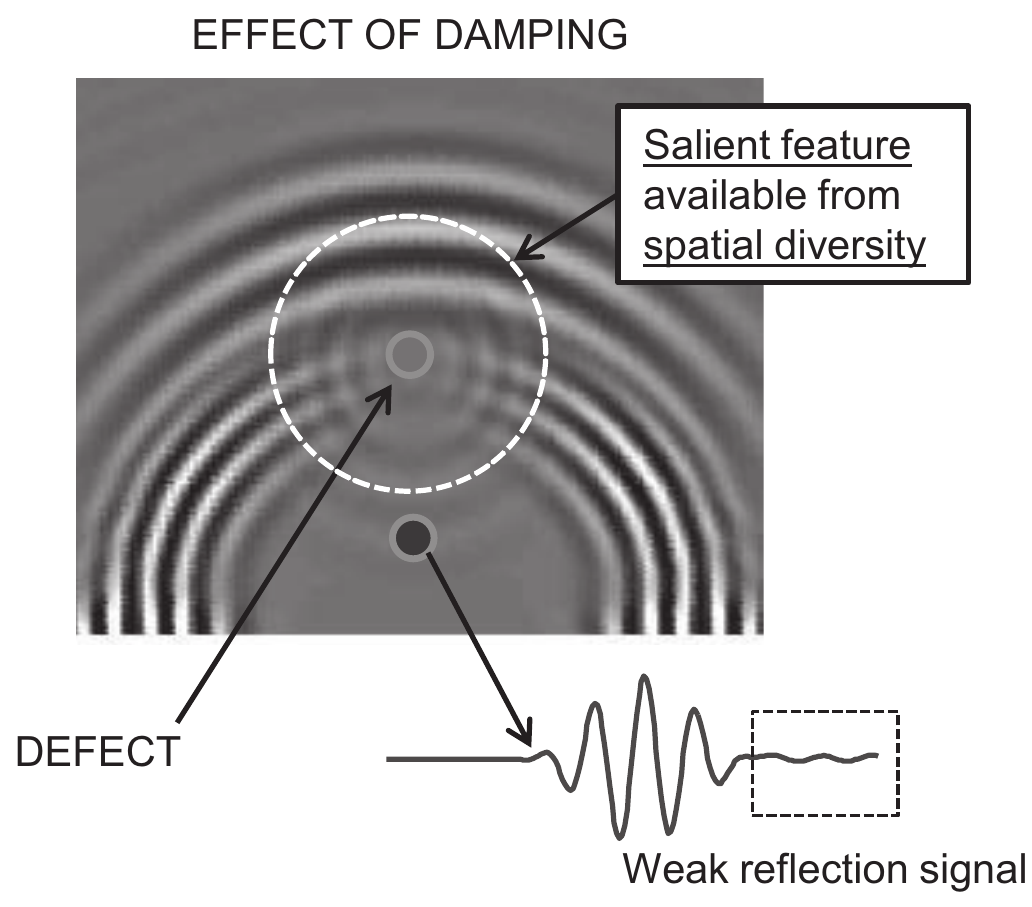}}
		\subfigure[Microstructural clutter]{\label{Effect_of_clutter} \includegraphics[scale=0.6]{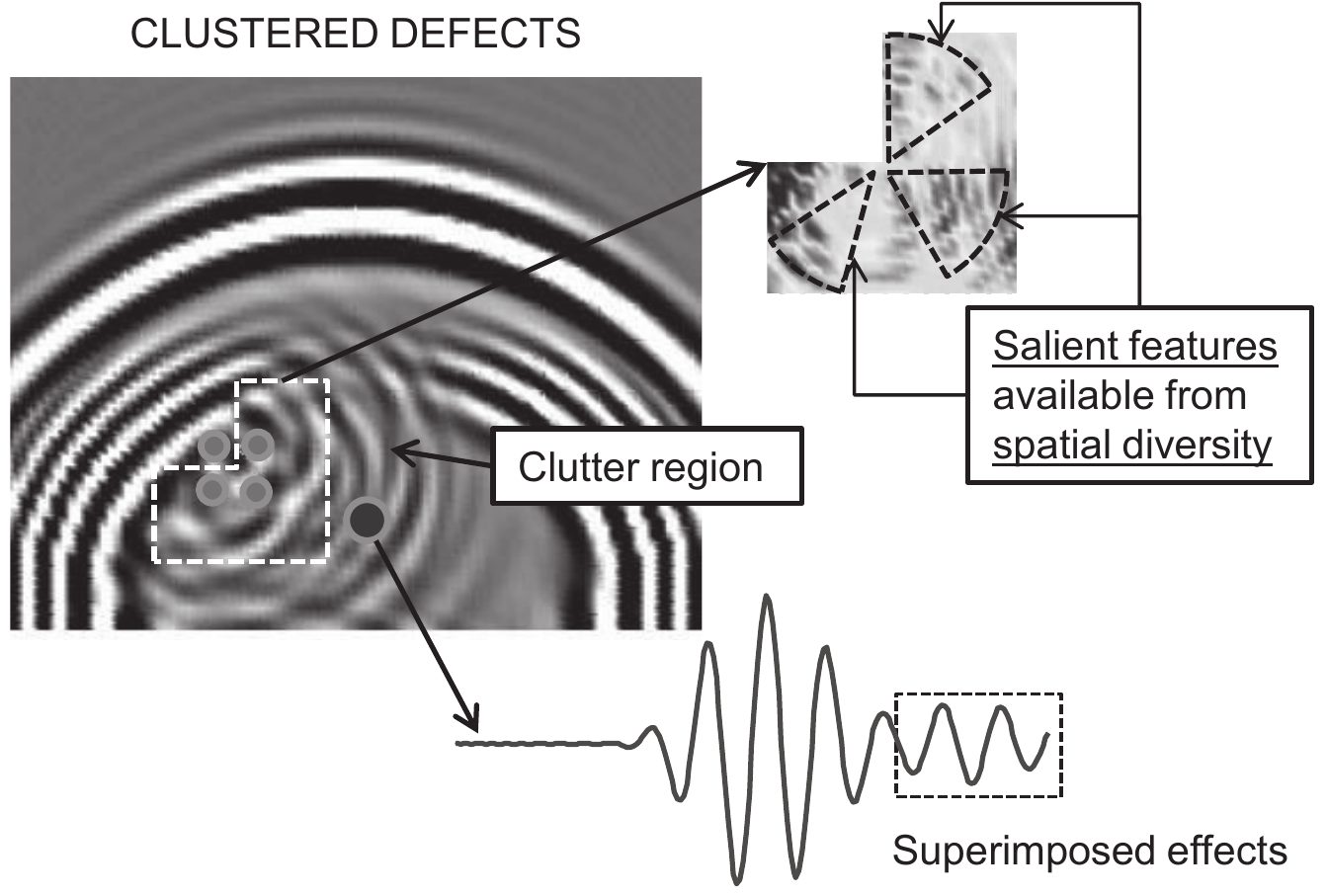}}
	\caption{\emph{Ambiguity observed in echo-pulse detection methods in the presence of material damping and microstructural clutter and opportunities for anomaly detection via spatial measurement diversity and identification of salient behavior}}
	\label{Issues}
\end{figure*}

Pitch-catch methods are at the core of \textit{in-situ} or \emph{online} structural health monitoring methodologies (SHM)~\cite{Kessler_Thesis,Kirikera_Balogun_Krishnaswamy_SHM}. In-situ SHM mandates that components are continuously monitored during their operational lives, and requires that appropriate actuation and sensing devices are embedded in the design. Embedding sensors in structures is an intriguing concept that has been explored with interesting results (e.g. for composite materials~\cite{Embedded_sensors_Chang}), although questions have been raised about the weakening effect of built-in devices on the material~\cite{Deborah_Chung_book}. The applicability of online monitoring techniques depends on a trade-off between the agility of these methods and the thoroughness of inspection that is achievable with methods involving full acoustic or optical access to the \emph{entire} structural domain.

Radar-based approaches work well under the assumption that the waves travel in the undamaged portion of the specimen without significant attenuation or dispersion; this assumption is valid for thin metal structures with limited damping, small number of widely-spaced defects, and negligible contribution to dispersion from the material microstructure.  However, radar based pitch-catch methods break down when these assumptions are relaxed. This is illustrated with a simple example in Fig.~\ref{Issues}, which depicts the results of a finite element simulation of S- and P-waves in a thin plate with soft inclusions. Figure~\ref{Effect_of_damping} depicts the effect of damping in a lossy material, where the amplitude decay can be so pronounced that the signal processed by the receiver is too weak to be detected or interpreted, especially in the case of noisy signals. Similarly, Fig.~\ref{Effect_of_clutter} makes use of the simple problem of scattering induced by four inclusions to illustrate the issues arising in the presence of multiple damage zones with high proximity. The proximity of the scatterers produces wave interference and significant clutter, with some inclusions becoming \textit{de facto} acoustically invisible to the traveling wave, and further, the signal recorded by the receiver is hard to decipher, as it is tainted by multiple reflections. \jdh{On the other hand, in either of these examples, full spatial reconstruction of the wavefield in the neighborhood of the defect, when available, allows a visual identification of the damage zone from the interpretation of the inherent spatial richness of the wavefield.}

The virtue of spatial measurement diversity, coupled with the superior reconstruction capabilities available using laser interferometers, have inspired a powerful class of laser-enabled inspection methodologies~\cite{Michaels_Ruzzene_Michaels_Ultrasonics_2010,Sharma_et_al_2005}. A Scanning Laser Doppler Vibrometer (SLDV) allows non-contact measurement of points belonging to a scanning grid, which enables full reconstruction of the propagating wavefield and provides abundant measurement diversity.  Subsequent data processing, for example using methods based on space-time Discrete Fourier Transform (DFT)~\cite{Chiu,Alleyne} or incident wave removal through wavenumber filtering~\cite{Ruzzene_SMS_2007}, have been proposed to improve visualization of propagating wavefields from SLDV data.

This work explores a new automated approach for inference of structural anomalies, which uses local (in space and time) low-rank modeling to describe typical wavefield behavior, and identifies (nominally rare) structural defects based on deviations from this model.  Our approach is based on fusing notions of \emph{saliency} \cite{Itti98, Itti01, Harel06, Liu07}, with low rank \jdh{plus (sparse) outlier} modeling \cite{Chandrasekaran11, CandesPCA, Outlier}, along the lines of several recent efforts examining these techniques in computer vision applications \cite{Yan10, Shen12}.  Saliency is often described in terms of human perception (e.g., in the study of visual saliency); here, our goal is to remove the human component in the diagnostic inference task in order to automate the inference procedure, in turn making it suitable for embedding in intelligent multi-step structural monitoring routines. The proposed approach is largely \emph{agnostic}, employing only minimal knowledge of the geometric and acoustic properties of the medium to perform the detection.

The remainder of this paper is organized as follows. In Section~\ref{Saliency informed anomaly detection} we state our material diagnostic problem in detail, specify our model for identifying saliency in the propagating wavefield data, and describe how this model is employed in our proposed automated inference approach.  Section~\ref{Applications and results} contains a validation of the proposed approach conducted using synthetic data in the context of two benchmark anomaly detection problems. We examine the robustness of our procedure applied to spatially downsampled data in Section~\ref{Discussion}.  Concluding remarks and future directions are discussed in Section~\ref{Conclusions and recommendations for future work}.

\section{Saliency based anomaly detection}\label{Saliency informed anomaly detection}

The objective of \emph{anomaly detection} is to identify atypical patterns in data \cite{Patcha07, Chandola09}.  Here, our aim is to identify the locations (or regions) of a structure at which some structural anomalies exist, from measurements in the form of kinematic time histories gathered at a set of discrete points on the structure's surface when the structure is excited by a propagating acoustic wave.  Propagating wavefields, of course, exhibit characteristics that depend upon the material properties of the medium (e.g., elastic moduli and density) and the externally applied excitation (e.g., carrier frequency and bandwidth).  In addition, the interactions between the propagating wavefields and structural anomalies produce \emph{signatures} in the measured data (e.g., reflections from, or interactions with, localized regions with different material properties). Our approach here identifies and exploits subtle differences between these signatures and otherwise ``nominal'' wave propagation characteristics to infer the locations of structural anomalies.  For clarity of exposition we restrict our discussion in the sequel to two-dimensional plate-like structures, though our proposed approach may be generalized to three-dimensional structures.

\subsection{Domain discretization and region partition}

Wave motion in two dimensions corresponds to the existence of a non-zero spatiotemporally evolving field $\phi(x,y,t)$, where $x,y$ are coordinates of locations in the material domain, $t$ is time and $\phi$ is a certain physical descriptor relative to the type of wave being considered. In two-dimensional plate-like structures experiencing flexural wave propagation, for instance, $\phi(x,y,t)$ can be a single displacement component, say $w(x,y,t)$, directed along the direction normal to the plane of the domain. In our numerical simulations, as well as in experimental acquisitions involving a laser scan, $w(x,y,t)$ is replaced by an array of time histories of nodal degrees of freedom $\bm w(t)$ belonging to a grid. For the sake of exposition, let us assume for simplicity that the grid is rectangular and has dimensions $N_1 \times N_2$. We suppose that the wavefield acquisition times are also discretized, so that time histories may be obtained at each point for times indexed by the discrete indices $\tau=1,2,\dots,T$. As a result, the full spatiotemporal evolution of the system is expressed in the form of a $N_1 \times N_2 \times T$ \emph{data cube} comprising one length $T$ time history for each grid point.

\begin{figure*}[t]
\centering
		\subfigure[Criterion for time shift and windowing]{\label{Regions_sequence1} \includegraphics[scale=0.6]{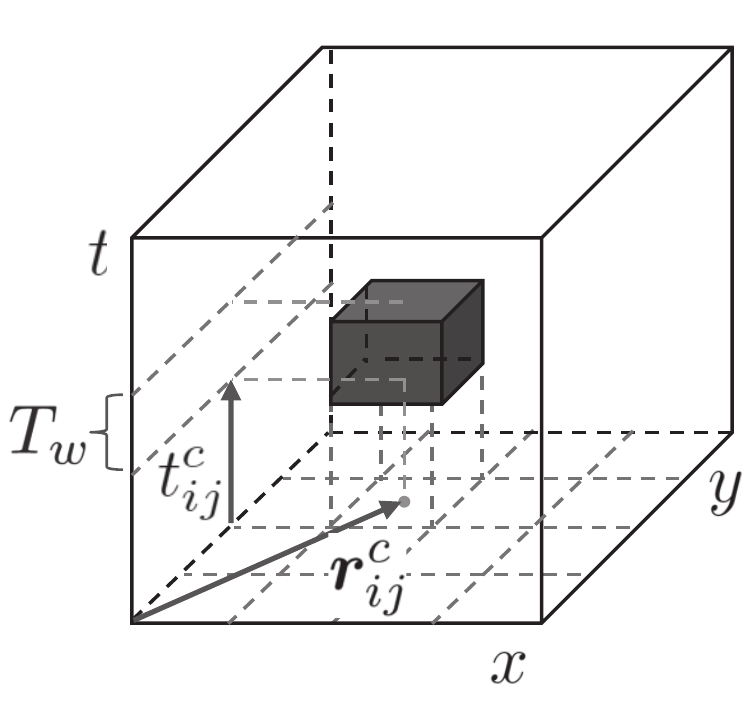}}
\hspace{0.4in}
		\subfigure[Selection of time shifting parameters]{\label{Regions_sequence2} \includegraphics[scale=0.6]{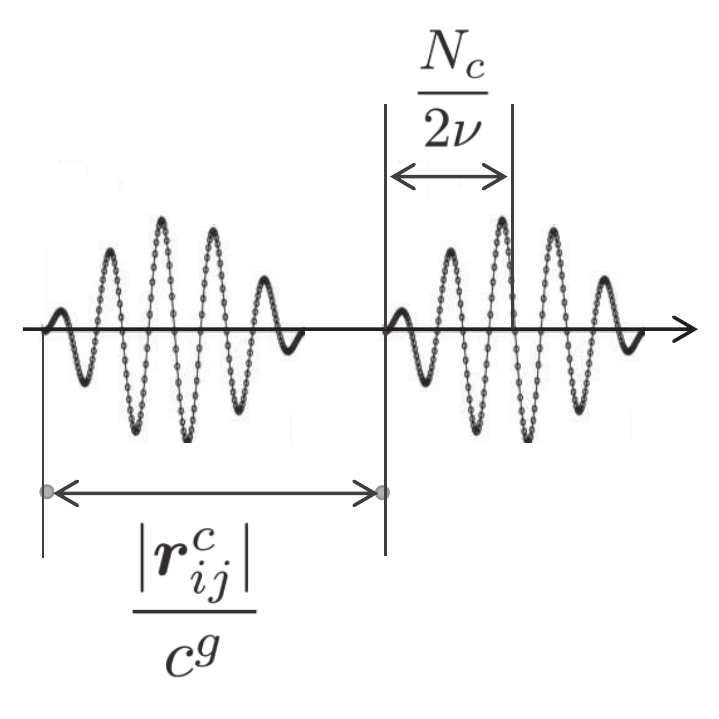}}
\hspace{0.4in}
		\subfigure[Schematic sequence of data cubes]{\label{Regions_sequence2} \includegraphics[scale=0.6]{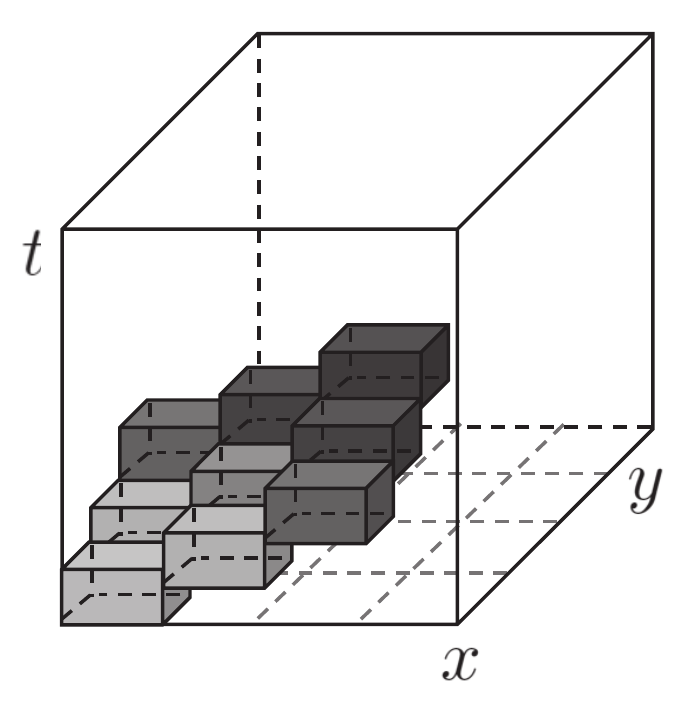}}
	\caption{\emph{Procedure for the construction of \jdh{$\widetilde{N}_1 \times \widetilde{N}_2$} region-wise data cubes \jdh{of size $p_1 \times p_2$} from the \jdh{$N_1 \times N_2$} space-time histories. \jdh{ Subfigure (c) highlights the role of time shifting in the identification of the regional data cubes. Every regional data cube is of size $p_1 \times p_2 \times T_w$, although the centering of each window in the time domain varies with the location of the corresponding region. The intensity of the regions color is proportional to the distance from the excitation point.}}}
	\label{Regions_windowing}
\end{figure*}

Conceptually, let us consider a partition of the spatial domain into $\widetilde{N}_1 \times \widetilde{N}_2$ rectangular (identical in shape and size) regions, where $\widetilde{N}_1<N_1$ and $\widetilde{N}_2<N_2$. Each region of the partition contains a subset of $p_1$ nodes along the $x$ direction, and $p_2$ along the $y$ direction. The precise relation between $\widetilde{N}_1, \widetilde{N}_2$ and $p_1, p_2$ depends on the partition (eg., whether two neighboring regions are allowed to share their boundary nodes or not). We assume here that each region shares its sides with the neighbors, in which case we have
\begin{equation}\label{centroid_tof}
p_1=\left(\frac{N_1-1}{\widetilde{N}_1}\right)+1, \,\,\,\,\,\,\, p_2=\left(\frac{N_2-1}{\widetilde{N}_2}\right)+1
\end{equation}
As customary in wave-based detection procedures, we suppose here that the structure is excited using an incident narrow-band burst applied at a single point in the domain. Following this assumption, it is possible to construct a spatiotemporal windowing of the displacement time histories informed by space and time characteristics of the excitation as follows.

Let $(i,j)\in\{1,2,\dots,\widetilde{N}_1\}\times \{1,2,\dots,\widetilde{N}_2\}$ index regions of the partition.  Inside the $(i,j)$ region, the expected time-of-flight of the centroid of the incident wave packet (i.e. the centroid of the envelope of the burst) to the physical centroid of the region is given by
\begin{equation}\label{centroid_tof}
t_{ij}^c = \frac{|\bm r_{ij}^c|}{c^g} + \frac{N_c}{2 \nu}
\end{equation}
where $c^g$ is the group velocity of the packet, $\bm r_{ij}^c$ is the position vector of the region centroid from the point of application of the excitation, $N_c$ is the number of cycles in the burst and $\nu$ is the wave frequency. The arrival time in \eqref{centroid_tof} can be seen as the sum of the arrival time of the front of the packet ($|\bm r_{ij}^c| / c^g$) plus the time interval corresponding to half burst ($N_c / 2 \nu$).
Now, for each sample location inside the selected region we retain only the first $T_w<T$ time samples immediately following the estimated incidence time. This results in the generation of a \emph{local} (region wise) data cube of size $p_1 \times p_2 \times T_w$. The procedure is schematically depicted in Fig.~\ref{Regions_windowing}.

This spatio-temporal ``preprocessing,'' essentially aligning the wave motion event at all locations, is a key component of our approach.  The windowing process is the only instance in the present formulation in which some knowledge of the wave propagation characteristics is invoked in the inference process. It is nevertheless worth noting that an estimate of group velocity does not necessarily require \emph{a priori} knowledge of the medium or construction of a model, as it can be inferred from a simple time of flight calculation, performed for instance between two virtual sensors placed in the domain along a direction where the likelihood of anomalous behavior is considered low (e.g. along one boundary).

\subsection{Anomaly localization as a sparse estimation task}\label{Anomaly localization as an estimation task}

Our defect or anomaly localization problem can be understood fundamentally as a \emph{sparse estimation} task.  Let us introduce a matrix $X\in\mathbb{R}^{\widetilde{N}_1\times \widetilde{N}_2}$, which we interpret as a \emph{feature matrix} whose entries correspond to distinct spatial regions of the partitioned domain.  Each element of the feature matrix corresponds to a contiguous region of the material domain.  Informally, we may view $X$ as a spatial ``map'' that indicates the locations of anomalies in the material.  Formally, let us suppose that entries of $X$ take the value $0$ at locations that correspond to anomaly-free homogenous regions, and are otherwise nonzero. We say $X$ is \emph{sparse} whenever most of its entries are zero; here, this corresponds to the number of regions of the material containing anomalies being small relative to the total number of regions.  We may consider the class of all sparse $X$ as elements of a \emph{feature space} ${\cal X}$.

The feature matrix $X$ is, of course, an abstraction, and we are not able to observe it directly.  Instead we must make inferences about $X$ by examining the response of the material to the incident traveling wavefield.  Our task is fundamentally an inverse problem, which we can formalize as follows.  Suppose that there exists an unknown (and potentially complicated) function $s$ that maps an element $X$ from the feature space ${\cal X}$ to a corresponding element of a \emph{structure space} $\mathcal{S}$. Elements $s(X)\in\mathcal{S}$ correspond to the real, physical material which exhibits the structural defects and inhomogeneities defined by their corresponding abstract feature matrices. We make measurements of $s(X)$ of the form of surface displacement time histories measured at locations on the material surface denoted by the ordered pair $(l,m)\in\{1,2,\dots,N_1\}\times\{1,2,\dots,N_2\}$.  We can model the observations as elements $y_{l,m}( s(X) )$ of an \emph{observation space} $\mathcal{Y}$. Note that each measurement is a function of some implicit parameters (e.g., wavelength and frequency) of the incident wave used for diagnostics, but for the sake of notational clarity we let the dependence on these wave parameters be implicit in the function $y$.

Given this framework, the detection task can be succinctly stated as follows: our aim is to infer $X$ (the sparse feature matrix, whose entries indicate the presence of structural anomalies) from a set of measurements $y_{l,m}( s(X) )$, obtained at a set of locations described by elements $(l,m)\in\{1,2,\dots,N_1\}\times\{1,2,\dots,N_2\}$.  Our proposed inference approach is based on identifying salient features in the propagating wavefield; that is, identifying local deviations from shared typical behavior exhibited by the bulk of the material.

\subsection{Low rank and sparse models for spatiotemporal saliency}

The essential premise behind \emph{saliency-informed diagnostics} is the notion that, in every region of the domain that is sufficiently far from any defects, the displacement time histories will exhibit some similar, but unknown, ``typical'' behavior, while the time histories recorded in spatial regions in the immediate vicinity of a defect will exhibit some (also unknown) signature of the defect that is different from the typical response observed in the bulk of the domain. The regions exhibiting atypical behavior are referred to here as \emph{salient}.  When only a few regions exhibit this atypical behavior, this notion of \emph{saliency} can be viewed as a generalization of the concept of \emph{sparsity}, which has been a central theme in signal processing, statistics, and machine learning research in recent years (see, for example, the voluminous works online at \cite{Web_RiceCS} which build upon the initial contributions in \emph{compressed sensing} \cite{Candes1, Donoho, Candes2, HauptRP, Dantzig}).

Our aim here is to identify a function of the measured data that serves as a surrogate, or \emph{proxy}, for the unknown (sparse) feature matrix $X$.
%
Recall that, following the time-windowing (described above) our measured data comprises an $N_1\times N_2 \times T_w$ data cube, where the data corresponding to each spatial \emph{region} is a $p_1\times p_2 \times T_w$ cube.  \jdh{Let us recall that all the time-windows have the same length $T_w$, although, because of shifting, they represent distinct actual time intervals.} For each time ``slice'' of the data cube (indexed by $\tau\in\{1,2,\dots,T_w\}$), we can interpret the data from each region as a $p_1\times p_2$  \jdh{two-dimensional} array or, \jdh{in equivalent vectorized form}, a vector of dimension $p_1p_2$.
Let $M_{\tau}(i,j)$ denote the $p_1 p_2$-dimensional vector associated with the $i,j$ location of the feature space, where $i,j\in\{1,2,\dots,\widetilde{N}_1\}\times\{1,2,\dots,\widetilde{N}_2\}$. Now, we assume that at each time step $\tau$ there exists a common $r$-dimensional linear subspace $\mathcal{U}_{\tau}\subset \mathbb{R}^{p_1 p_2}$, with $r < p_1 p_2$, that well approximates the spatiotemporal data inside the anomaly-free regions.  Let $P_{\mathcal{U}_{\tau}}$ denote the orthogonal projection onto the subspace ${\mathcal{U}_{\tau}}$ and consider the matrix $Z_{\tau}\in\mathbb{R}^{\widetilde{N}_1 \widetilde{N}_2}$ whose $(i,j)$-th entry is
\begin{equation}\label{eqn:proxy}
Z_{\tau}(i,j) = \|P_{\mathcal{U}_{\tau}} M_{\tau}(i,j) - M_{\tau}(i,j)\|_2^2,
\end{equation}
for $i=1,2,\dots, \widetilde{N}_1, \ \ j=1,2,\dots,\widetilde{N}_2$, where $\|\cdot\|^2_2$ is the square of the standard Euclidean norm.
When an entry of $Z_{\tau}$ is significantly different from zero, it means that the local wavefield at the corresponding location in the domain has significant energy outside of the subspace $\mathcal{U}_{\tau}$ (or, equivalently, is not well-represented as a linear combination of vectors in $\mathcal{U}_{\tau}$).  In this interpretation, the matrix $Z_{\tau}$ serves as a proxy for the true (unknown) feature matrix $X$, which identifies the locations whose windowed time histories deviate significantly from the bulk or typical behavior according to the selected linear subspace model.


Note that the model we employ here can be interpreted in terms of a \emph{low-rank plus outlier} data model.  In particular, let us denote by $M_{\tau}$ the $p_1 p_2 \times\widetilde{N}_1\widetilde{N}_2$ matrix formed by assembling the vectorized data from each region at time step $\tau$.  Our approach is equivalent to decomposing $M_{\tau}$ as
\begin{equation}\label{eqn:LRPS}
M_{\tau} \approx L_{\tau} + C_{\tau} 
\end{equation}
where $L_{\tau}\in\mathbb{R}^{p_1 p_2 \times\widetilde{N}_1\widetilde{N}_2}$ is a low-rank matrix and $C_{\tau}\in\mathbb{R}^{p_1 p_2 \times\widetilde{N}_1\widetilde{N}_2}$ is ``column-sparse'', having only a few non-zero columns, which can be interpreted as vectors lying outside of the subspace spanned by the columns of $L_{\tau}$. 
 Identifying the ``outlier'' vectors, which correspond to the non-zero columns of $C_{\tau}$, is the aim of so-called Robust Principal Component Analysis (or Robust PCA), and many procedures have been examined to address this problem~\cite{Gnan72, Fischler81, DeLaTorre03, Ke05}, including recently-proposed convex formulations~\cite{Chandrasekaran11, CandesPCA, Outlier}.

\begin{figure}[h]
\centering
		\includegraphics[scale=0.45]{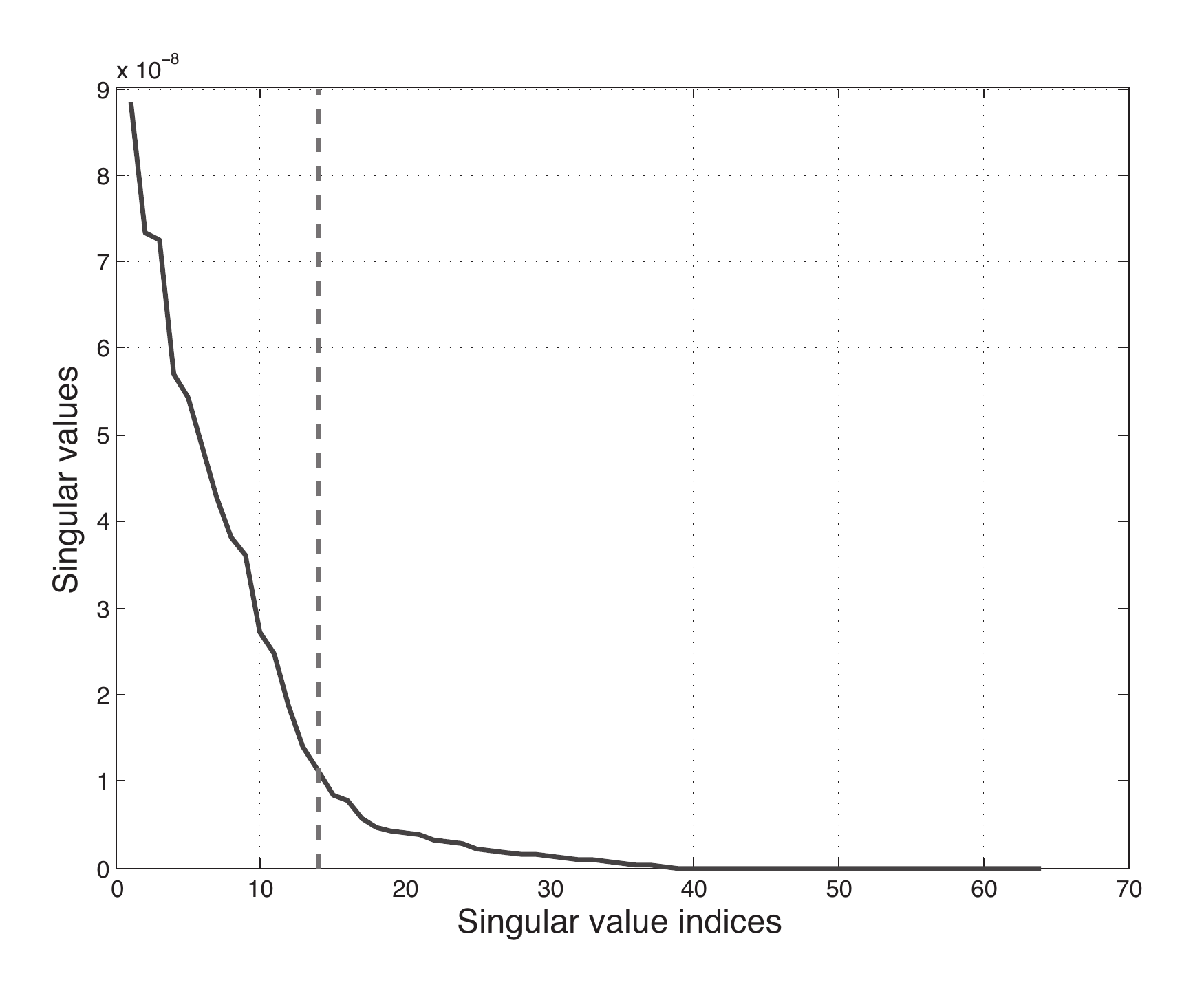}
	\caption{\emph{Knee in the singular value curve}}
	\label{SVD_Knee}
\end{figure}

Here, for the sake of simplicity, we perform the decomposition of $M_{\tau}$ using an approach based on standard matrix operations. Specifically, we identify a rank-$r$ approximation $\widehat{L}_{\tau}$ of $M_{\tau}$ as a solution to the optimization problem
\begin{equation}\label{eqn:low-rank-SVD}
\widehat{L}_{\tau} = \min_{M\in\mathbb{R}^{p_1 p_2 \times \widetilde{N}_1\widetilde{N}_2}: \mbox{rank}(M) \leq r} \|M_{\tau} - M\|_F,
\end{equation}
where the notation $\|\cdot\|_F$ corresponds to the matrix Frobenius norm.  The Eckart-Young Theorem establishes that the solution $\widehat{L}_{\tau}$ of \eqref{eqn:low-rank-SVD} is easily obtained via the \emph{singular value decomposition} (SVD) \cite{Eckart36, Stewart93}. Write the singular value decomposition of $M_{\tau}$ as
\begin{equation}\label{SVD_Calculation}
M_{\tau}=USV^H,
\end{equation}
where $U$ and $V$ are unitary, $S$ is a rectangular diagonal (and nonnegative) matrix whose diagonal entries are the singular values ordered from largest to smallest, and the superscript $H$ denotes the matrix Hermitian (complex conjugate transpose) operation. Then, the solution $\widehat{L}_{\tau}$ of \eqref{eqn:low-rank-SVD} is obtained as
\begin{equation}\label{SVD_Procedure}
\widehat{L}_{\tau}=US_rV^H,
\end{equation}
where $S_r$ is obtained from $S$ by keeping only the $r$ largest singular values and zeroing the rest. Note that the first $r$ columns of $U$ comprise an orthonormal basis for the common subspace.  In the context of \eqref{eqn:proxy} above, we may let $U_r$ be the $p_1 p_2 \times r$ matrix comprised of only the first $r$ columns of $U$, then $P_{\cal U} = U_r(U_r^TU_r)^{-1}U_r^T = U_r U_r^T$.  In practice, the dimension $r$ of the common subspace is inferred empirically from the \emph{knee} of the plot of (ordered) singular values, which corresponds to the point beyond which the decay in the singular values becomes gradual (see Fig.~\ref{SVD_Knee}).

From an implementation standpoint, our inference approach proceeds as follows.  At each time step, we construct  the low rank approximation $\widehat{L}_{\tau}$ and a corresponding estimate of the column-sparse matrix $\widehat{C}_{\tau} = M_{\tau} - \widehat{L}_{\tau}$, and we identify the columns of $\widehat{C}_{\tau}$ that have non-negligible energy.  The significant columns correspond to salient regions in the domain, i.e., regions with high likelihood to contain an anomaly. These estimates are finally aggregated over the $T_w$ distinct time steps of the considered time window to account for the complete scattered wavefield induced by each anomaly. The identification of the significant columns operation inevitably involves the definition of appropriate thresholding levels which can slightly vary from case to case. Additional implementation details are provided in the next section in the context of specific applications.

\section{Application to two-dimensional transversal wave propagation} \label{Applications and results}

The approach presented above is now tested in a simulated environment using data generated via the finite element method (FEM). The application selected to test the method is the benchmark problem of circular-crested transversal waves excited in a thin plate via application of an out-of-plane point force at one node. From the point of view of the wavefield topology, this scenario is representative of both flexural plate waves and guided Lamb waves, therefore the following results and conclusions about the efficacy of the anomaly detection algorithm will be valid for both. In this work, flexural waves are modeled according to Mindlin's plate theory, which involves three \jdh{nodal} degrees of freedom, one out-of-plane displacement $w(x,y,t)$ and two rotations $\theta_x(x,y,t)$ and $\theta_y(x,y,t)$, although the anomaly detection analysis will be restricted to the deflection $w(x,y,t)$ which is, from the point of view of an equivalent laser-based acquisition, the degree of freedom which is more directly measurable from a surface scan.

The choice of flexural waves is here primarily dictated by the relative simplicity of the corresponding FEM model and the advantages (in terms of ambiguity removal) resulting from dealing with a single-mode wave solution. The excitation is applied at one corner of the plate in order to maximize the radius of the propagation domain for a given mesh size. The time history of the load is a $5$-cycle narrow band burst, as customary in ultrasonic testing. The domain is taken to be square ($L_x=L_y=L$), and the dimensions are selected jointly with the material properties and the carrier frequency of the excitation such that $L/ \lambda=25$, where $\lambda$ is the carrier wavelength of the induced wave. \jdh{Note that the analysis is conducted here in fully non-dimensional form to emphasize the scalability of the results, which is an important feature of this treatment. However, for the sake of completeness, we list the geometric and material properties used in the simulations: material properties (Aluminum, $E=71 \textrm{GPa}, \nu=0.33, \rho=2700\textrm{Kg}/\textrm{m}^3$); dimensions ($0.25 \textrm{m} \times 0.25 \textrm{m}$); thickness ($0.005 \textrm{m}$). The excitation has a carrier frequency of $500  \textrm{kHz}$, which corresponds to a wavefield with carrier wavelength of $0.01 \textrm{m}$.} Exploiting the square shape, a partition in $N \times N$ square regions of identical size is adopted in the detection process. A structured mesh comprising $256 \times 256$ square elements is used to generate an accurate solution free of unwanted numerics-induced dispersive and directional effects. This data field can be \emph{a posteriori} coarsened via under-sampling of the nodal solution to explore the robustness of the method against coarse data acquisition.

\subsection{Benchmark problem $\sharp 1$. Scattered defects}\label{Scattered anomalies}

\begin{figure*}[t]
\centering
		\subfigure[Final snapshot of wave motion]{\label{Def_3_Regions_8_8_Wavefield} \includegraphics[scale=0.6]{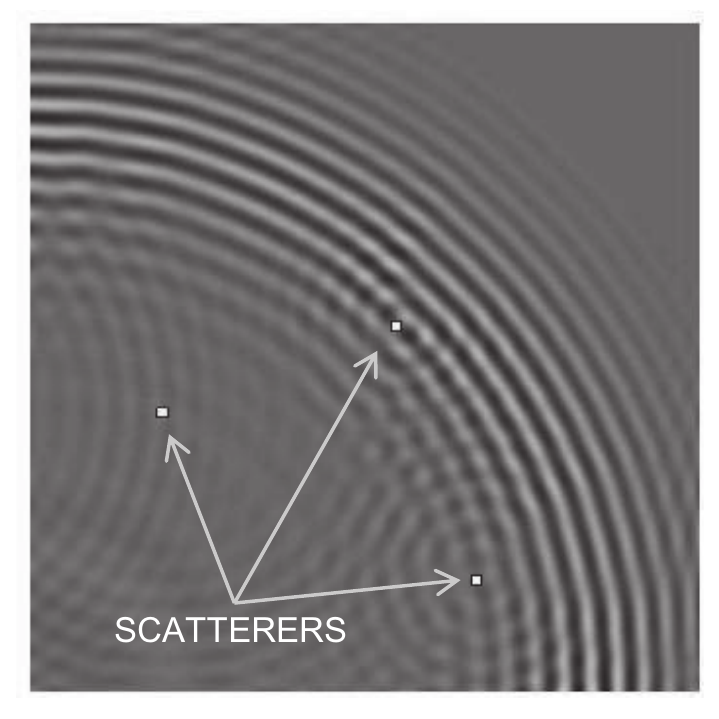}}
\hspace{0.25in}
		\subfigure[Saliency map]{\label{Def_3_Regions_8_8_Saliency_Map} \includegraphics[scale=0.55]{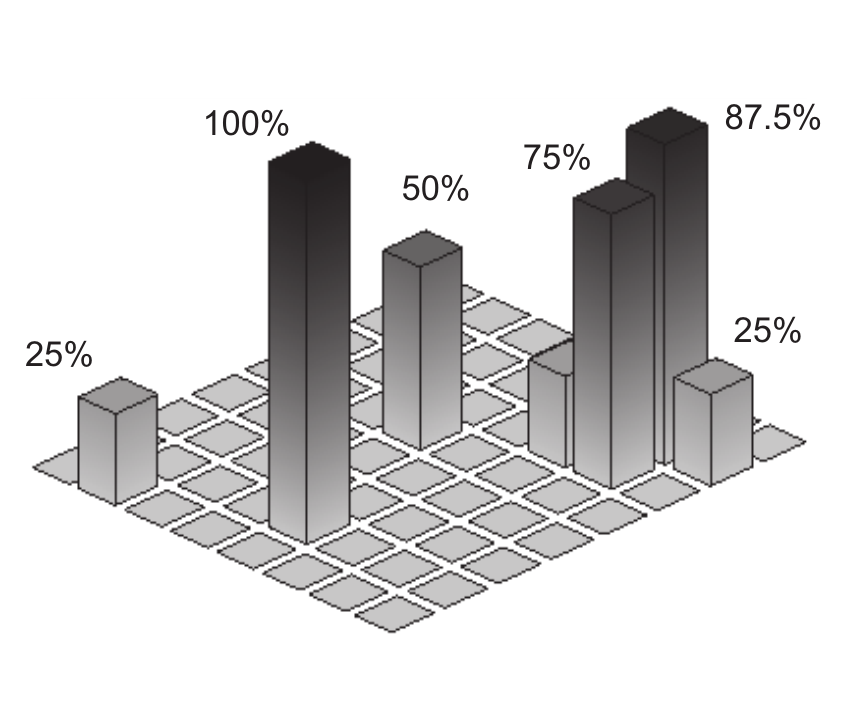}} 
\hspace{0.25in}
    	\subfigure[Identification of salient regions]{\label{Def_3_Regions_8_8_Outcome} \includegraphics[scale=0.6]{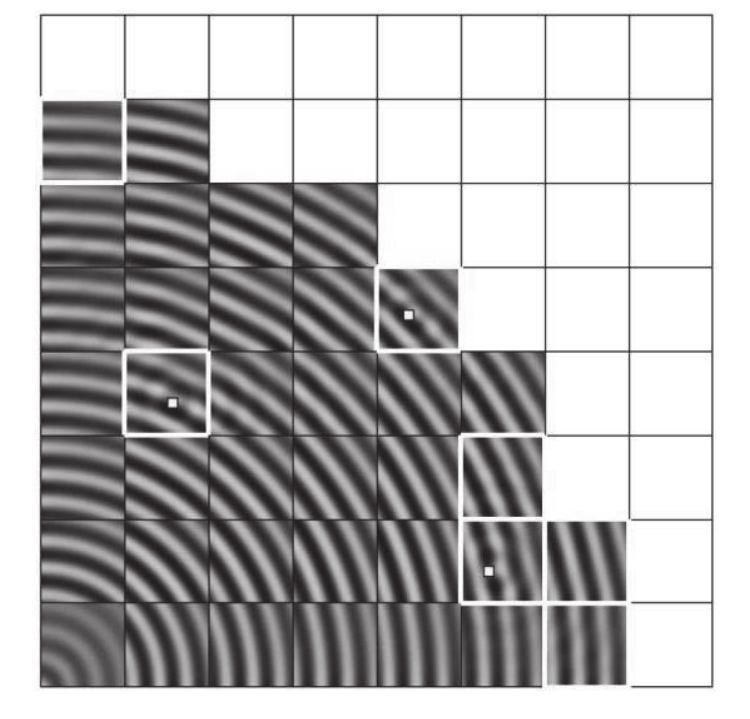}}
	\caption{\emph{Anomaly detection via saliency map constructed using coarse $8 \times 8$ region partition. At this level of refinement, the method over-predicts the salient regions, thus providing a moderately accurate but conservative estimate.}}
	\label{Def_3_Regions_8_8}
\end{figure*}

Three anomalies are scattered in the domain at locations $x_1 = 0.2 L \ , \ y_1 = 0.42 L \ , \ x_2 = 0.55 L \ , \ y_2 = 0.55 L \ , \ x_3 = 0.67 L \ , \ y_3 = 0.17 L$, respectively. The anomalies are introduced by changing the material properties of each finite element that contains an anomaly point. In this example, the Young's modulus and density are increased inside the defected element by two orders of magnitude to simulate a distribution of stiff inclusions in the material. \jdh{Let us recall that the material is homogeneous and isotropic and in its pristine state other than at the locations where the anomalies are explicitly introduced.}  A snapshot of the computed wavefield in the presence of defects is shown in Fig.~\ref{Def_3_Regions_8_8_Wavefield}. The saliency analysis is first implemented using a coarse $8 \times 8$ partition of the domain, as shown in Fig.~\ref{Def_3_Regions_8_8_Outcome}, such that each region extends over a $32 \times 32$-element mesh. By comparison of the wavefields in Fig.s~\ref{Def_3_Regions_8_8_Wavefield} and~\ref{Def_3_Regions_8_8_Outcome}, it can be noticed how, as a result of the region-wise time shifting performed according to \eqref{centroid_tof}, each regional local wavefield corresponds to the time instant at which the wave front impinges on the centroid of the region. This shifting operation \emph{de facto} replicates the wave ``state'' from one region to the next. As a result, the \emph{intrinsic} (not associated with the presence of defect) topological differences between regions are limited to two factors: \emph{a)} A phase shift (observable by comparing the wave crests) due to the difference between phase and group velocity ($c^g=2 c^p$ according to plate theory); \emph{b)} An increase in radius of curvature due to the time evolution of the radius of the circular-crested wave front ($R = c^g t$). As a result, the windowing procedure highlights the salient features associated with \emph{extrinsic} mechanisms (in this case the scattering induced by the inclusions, where this is observed), thus minimizing the ambiguity between real and spurious sources of salient behavior in the detection process. The white regions in Fig.~\ref{Def_3_Regions_8_8_Outcome} represent peripheral portions of the domain where the wave has not propagated within the simulated time window; the wavefield is there identically zero, and the method is naturally incapable of identifying any anomalies in those sectors.  \jdh{It is also worth pointing out that, while the anomalies are interpreted in this example as defects, they can represent any structural feature that acts  as a source of deviation from the nominal behavior of the plate, including fastener holes, thickness changes, etc. In fact, at this stage of development, the method would not be able to discriminate between these different sources of saliency.}

The time histories collected at the nodes of each $(i,j)$-th region are used to construct the $M_{\tau}$ matrices to feed to the optimization problem of  \eqref{eqn:low-rank-SVD}, where $\tau=1,2,\dots,T_w$ and here $T_w= 11$, i.e. $11$ time instants following the average time of flight $t^c_{ij}$ are considered. The low rank approximation $\widehat{L}_{\tau} $ is obtained by selecting $r = 14$ from inspection of the knee plot of Fig.~\ref{SVD_Knee}. With the information from each region at each time instant, a saliency map is constructed as shown in Fig.~\ref{Def_3_Regions_8_8_Saliency_Map} by counting how many times the energy of the corresponding column of the outlier estimate $\widehat{C}_{\tau} = M_{\tau} - \widehat{L}_{\tau}$ is sufficiently large (here we identify all columns of $\widehat{C}_{\tau}$ whose energy exceeds $25 \%$ of the largest outlier energy found in any region at the same time snapshot). The saliency map tells us how consistently each region is classified as \emph{salient} (with respect to the surrounding ones) over the whole time window of interest. For instance, if a region is given a value $50\%$ in the saliency map, that region has been found to be salient $50\%$ of the time instants. By highlighting the salient regions on the partitioned domain, the results can be verified against the known positions of the scatterers, as shown in Fig.~\ref{Def_3_Regions_8_8_Outcome}.

\begin{figure*}[t]
\centering
		\subfigure[$16 \times 16$ regions]{\label{Def_3_Regions_16_16_Outcome} \includegraphics[scale=0.75]{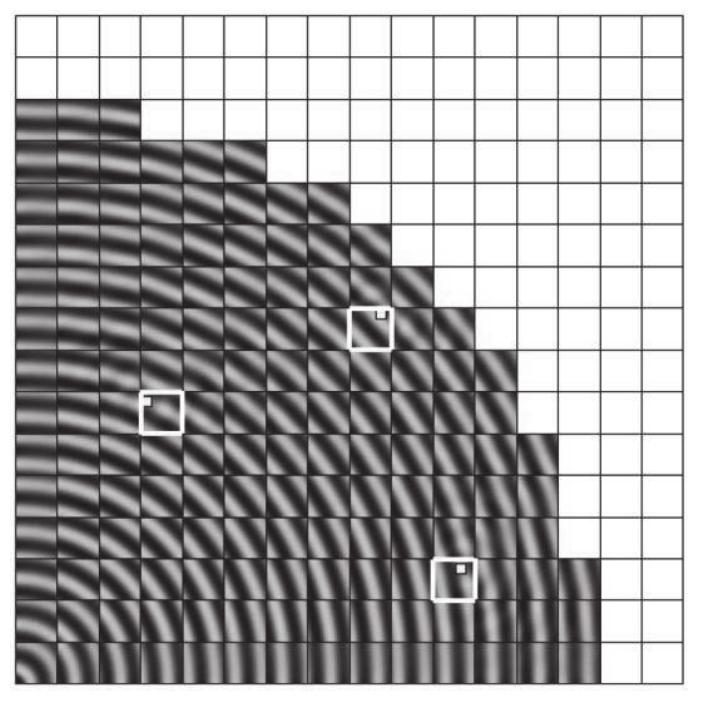}}
        \hspace{1in}
		\subfigure[$32 \times 32$ regions]{\label{Def_3_Regions_32_32_Outcome} \includegraphics[scale=0.75]{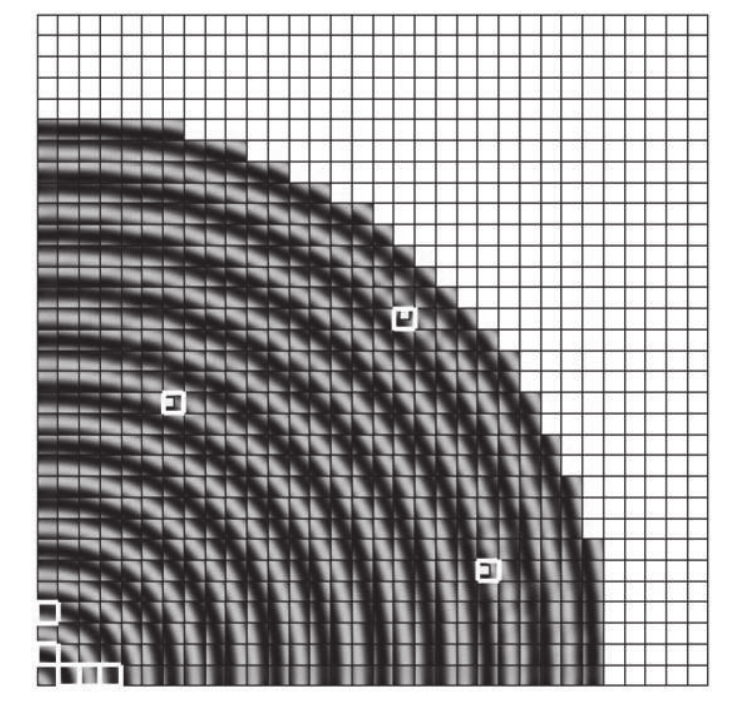}}
	\caption{\emph{Saliency analysis results via refinement of region partition}}
	\label{Def_3_Regions_Refinement}
\end{figure*}

It can be noted that, at this level of refinement, the method over-predicts the salient regions, thus providing a moderately accurate but conservative estimate. Let us now explore the sensitivity of the method to refinements of the region partition by considering first $16 \times 16$ and subsequently $32 \times 32$ regions, as shown in Fig.~\ref{Def_3_Regions_16_16_Outcome} and Fig.~\ref{Def_3_Regions_32_32_Outcome}, respectively. In the first case, the saliency analysis pinpoints accurately the defects, the outliers are avoided and the accuracy in defect localization has grown at the rate of the partition refinement. Reasons for the improved performance with respect to the $8 \times 8$ case are to be found in the fact that smaller regions allow focusing on limited wavefields that are less likely to be affected by spurious effects (such as boundary reflections, or reflections from the surrounding scatterers) that could be originating in neighboring regions. \jdh{Another explanation is that smaller regions are accompanied by tighter and more localized time referencing, which allows pinpointing the instances of scattering formation more precisely.}

The performance of the $32 \times 32$ case offers some interesting insight into the details of the outlier modeling philosophy. The defects are again found and localized very precisely, and the small size of the regions almost allows a determination of their point locations. However, a few outliers appear in the neighborhood of the excitation source. An explanation of this can be that the wavefield data structure inside small regions is very sensitive to differences in the curvature of the wave crests: the pronounced curvature observed in the regions surrounding the origin provides them with an element of saliency that competes with the one due to the scattering mechanisms. It is important to report that the neighborhood of the origin is found to be critical in many simulations, which suggests modest improvements of the algorithm geared toward eliminating this weakness.


\subsection{Benchmark problem $\sharp 2$. Line defect}\label{Line defect}

\begin{figure*}[t]
\centering
	   \subfigure[$8 \times 8$ regions]{\label{Def_Crack_8_8_Outcome} \includegraphics[scale=0.6]{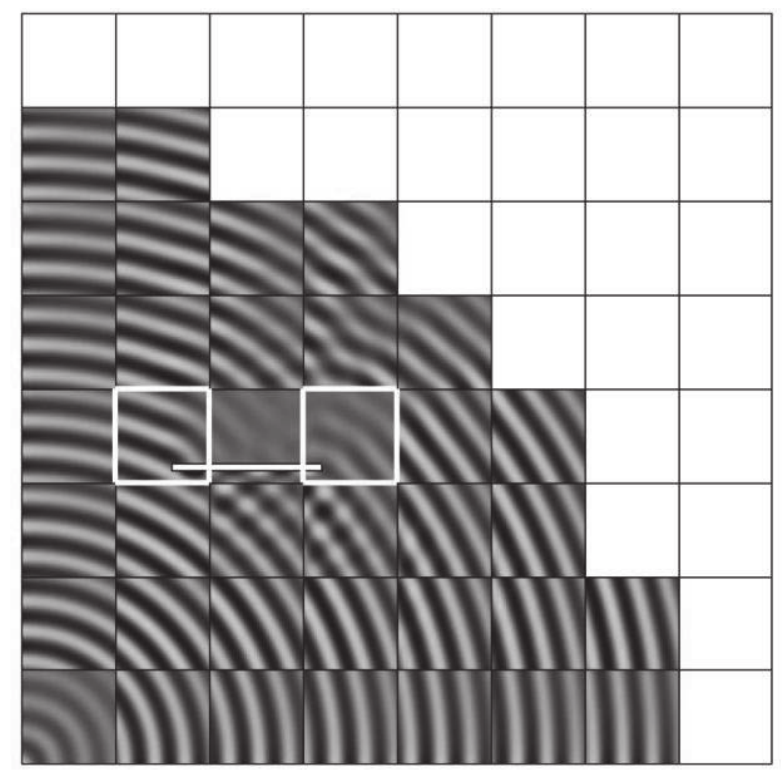}}
\hspace{0.25in}
		\subfigure[$16 \times 16$ regions]{\label{Def_Crack_16_16_Outcome} \includegraphics[scale=0.6]{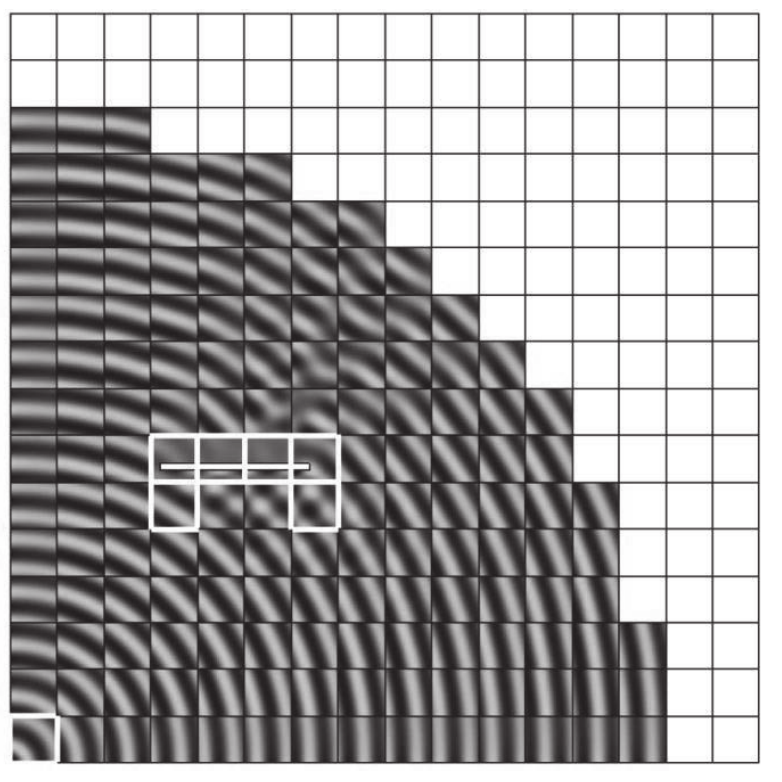}}
\hspace{0.25in}
    	\subfigure[$32 \times 32$ regions]{\label{Def_Crack_32_32_Outcome} \includegraphics[scale=0.6]{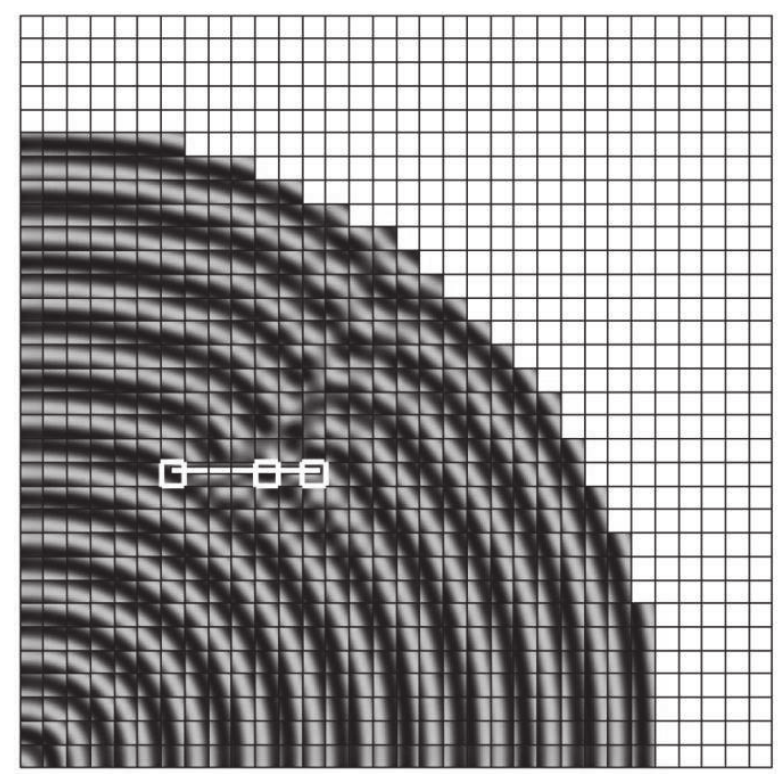}}
	\caption{\emph{Saliency analysis results via refinement of region partition - Line defect.}}
	\label{Crack_Regions_Refinement}
\end{figure*}

Let us test the method against the case of line defect, i.e. a region of anomaly extending over a considerable portion of the domain that overlaps with multiple region partitions at any stage of refinement. These conditions can be representative of a crack, or a region of de-bonding between material phases, or a path of void coalescence (which typically precedes the onset of crack). The defect is simulated by reducing by several orders of magnitude the Young's modulus and density of the material inside a one-element thick layer extending from $x_1 = 0.2 L$ to $x_1 = 0.4 L$. The results of the analysis are shown in Fig.~\ref{Crack_Regions_Refinement} for increasing levels of refinement. With $8 \times 8$ regions (Fig.~\ref{Def_Crack_8_8_Outcome}), the method does a good job at coarsely bounding the region of the defect. The saliency criteria embedded in the algorithm emphasize the regions surrounding the tips of the defect, as this is where the major local perturbations of the wavefield are observed in terms of perturbation of the curvature of the wave crests. The $16 \times 16$ refinement (Fig.~\ref{Def_Crack_16_16_Outcome}) allows complete localization of the defect path, but slightly over-predicts in the tip areas, and features one outlier by the origin, both effects being traceable back to the considerations on the crest curvature made above. The $32 \times 32$ refinement (Fig.~\ref{Def_Crack_32_32_Outcome}) interestingly eliminates the ambiguity in the neighborhood of the origin, captures and refines the localization of the defect tips and provides a spotty reconstruction of the defect path. \jdh{Note that the behavior of the method in the neighborhood of the source shows marked differences between the results of Fig.~\ref{Def_3_Regions_Refinement} and those of Fig.~\ref{Crack_Regions_Refinement}, for the same excitation conditions. This is an intrinsic feature of saliency-based detection methods, due to the very essence of saliency, which is not a property of a region, but a property of the region with respect to its neighbors. The same regional data are deemed salient or not relative to the salient features displayed by other regions in the domain. In this case, the energy available outside of the common subspace in the regions surrounding the excitation is comparable to that of the regions containing the scattered anomalies, but considerably less than that of the regions containing the crack.}

\section{Assessment of the method flexibility} \label{Discussion}

\subsection{Spatially deterministic and  random sub-sampling}\label{Subsampling}

The results presented so far rely on dense sampling ($p_1p_2$ nodal points within each of the $\widetilde{N}_1 \times \widetilde{N}_2$ regions). In this section we explore the limitations of the saliency based anomaly detection strategy involving a subset of the points from each region. The benefits of this kind of approach would be in dealing with experimentally acquired wavefields, where acquiring and processing less data would enable a faster acquisition phase, thus increasing the efficiency of the entire anomaly detection process. Coarser data are here easily obtained from FEM computed wavefields by sub-sampling the data in space.

\begin{figure*}[t]
\centering
	   \subfigure[Random sub-sampling ($n_z \approx 20 \%$)]{\label{Subsampling_3_Def_16_16_Rand_20} \includegraphics[scale=0.64]{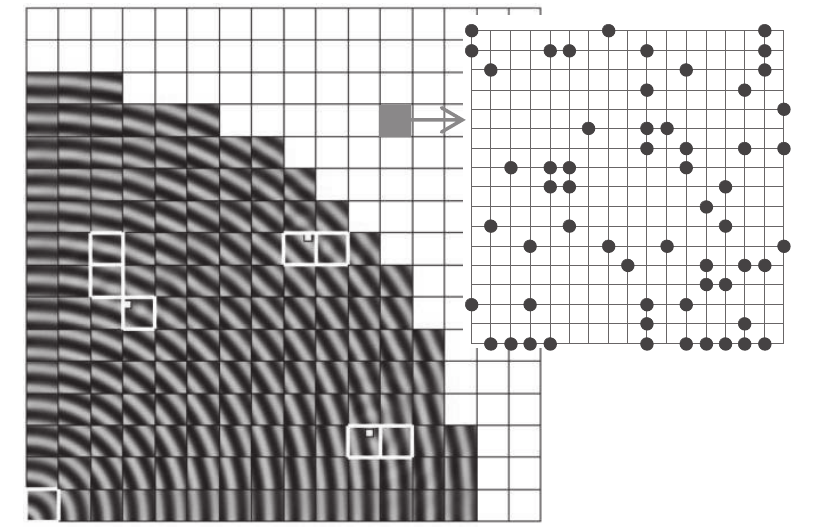}}
	   \subfigure[Random sub-sampling ($n_z \approx 10 \%$)]{\label{Subsampling_3_Def_16_16_Rand_10} \includegraphics[scale=0.64]{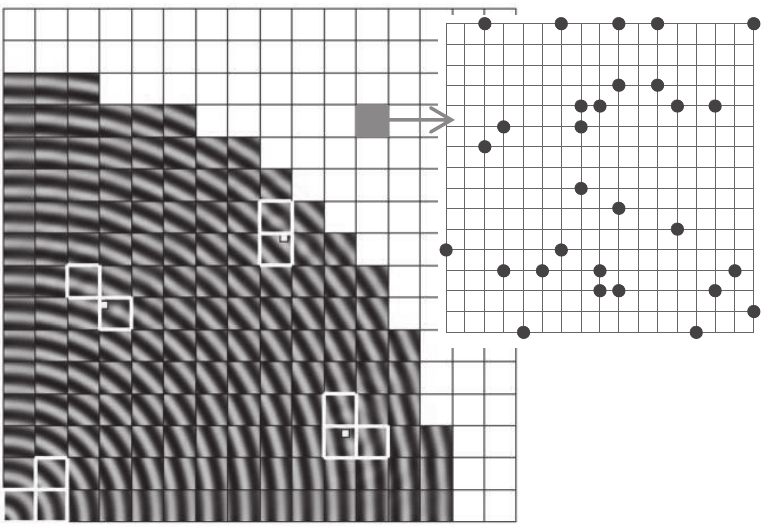}}
	   \subfigure[Random sub-sampling ($n_z \approx 7 \%$)]{\label{Subsampling_3_Def_16_16_Rand_7} \includegraphics[scale=0.64]{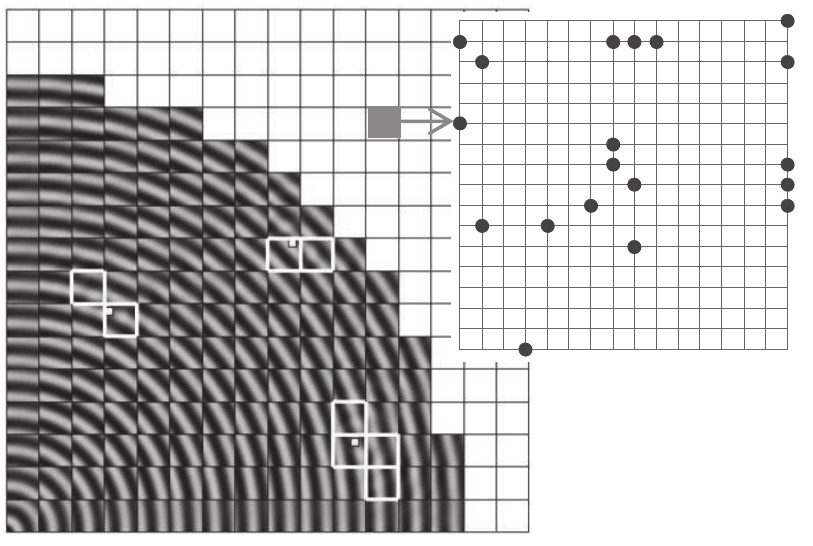}}
        \vspace{0.01in}
	   \subfigure[Deterministic double cross sub-sampling ($n_z \approx 22 \%$)]{\label{Full_Cross_Sub_sampling_3_Def_16_16} \includegraphics[scale=0.64]{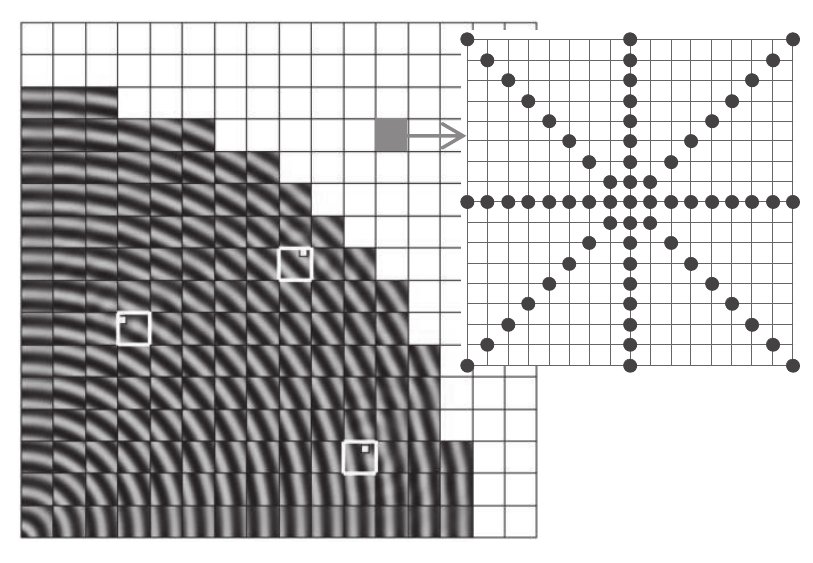}}
	   \subfigure[Deterministic double cross sub-sampling ($n_z \approx 11 \%$)]{\label{Full_Cross_Sub_sub_sampling_3_Def_16_16} \includegraphics[scale=0.64]{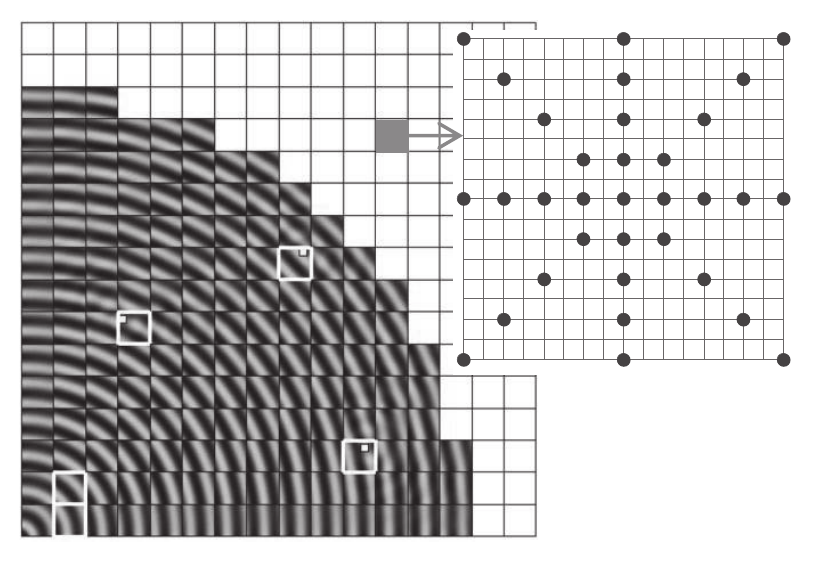}}
	   \subfigure[Deterministic double cross sub-sampling ($n_z \approx 6 \%$)]{\label{Full_Cross_Sub_sub_sub_sampling_3_Def_16_16} \includegraphics[scale=0.64]{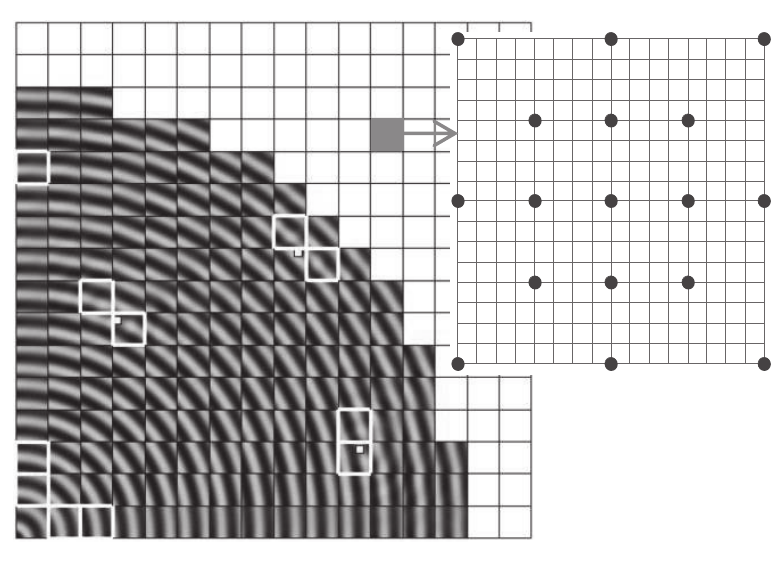}}
	\caption{\emph{Effect of subsampling with random and deterministic patterns - Scattered defects.}}
	\label{Subsampling_Effects_16_16}
\end{figure*}

We explore two main classes of sub-sampling techniques.  The first three cases (Fig.~\ref{Subsampling_3_Def_16_16_Rand_20},~\ref{Subsampling_3_Def_16_16_Rand_10} and~\ref{Subsampling_3_Def_16_16_Rand_7}) depict the results obtained using a random sub-sampling approach, with spatial sub-sampling ratios $n_z \approx 20 \%$, $n_z \approx 10 \%$ and $n_z \approx 7 \%$, respectively.  As expected, the performance degrades relative to the reference exhaustive case as $n_z$ decreases. However, in all cases depicted in the figure, the algorithm does manage to identify the correct \emph{neighborhoods} of the defects.  Based on the figures, the practical implications of this appear to be minor, as the regions identified using the sub-sampled data can be interpreted as slightly more conservative estimates of the actual defect locations.

However, it is important to note that random sub-sampling inevitably introduces some variability in the results.
We investigate this variability empirically here. We conduct $500$ independent trials (each corresponding to a different randomly generated subsampling pattern for each region) for each of several downsampling rates, and report the average number of regions corresponding to correct discoveries and false discoveries.  We also compute two ``regional'' error metrics that take into account whether the detections are within the immediate vicinity of the regions containing the true anomalies. These regional criteria are evaluated as follows.  If in a given trial, at least one of the discovered anomalous regions is an immediate neighbor of the true region containing the anomaly (or is the anomalous region itself), that anomaly is deemed discovered.  Similarly, false discoveries are only deemed false if they occur at least one region removed from any region containing a true anomaly. We also report the average number of false discoveries that occur in the $3\times 3$ region at the origin of the domain, motivated by the empirical observation that false discoveries often occur near the excitation source. The results are collected in Table~\ref{Table_subsampling}.

It can be noted how, while the performance decays as expected when the sub-sampling ratio $n_z$ decreases, the algorithm features quite remarkable detection capability even for very coarse acquisitions.
Further, the ``regional'' error metrics underscore the empirical observation above -- detections overwhelmingly occur in the immediate vicinity of the anomalous regions, and false discoveries (when present) are predominantly near the excitation source.  This suggests that simply disregarding these regions in the final estimate may be a useful, if simple, heuristic.\\

\begin{figure*}[t]
\centering
		\includegraphics[scale=0.7]{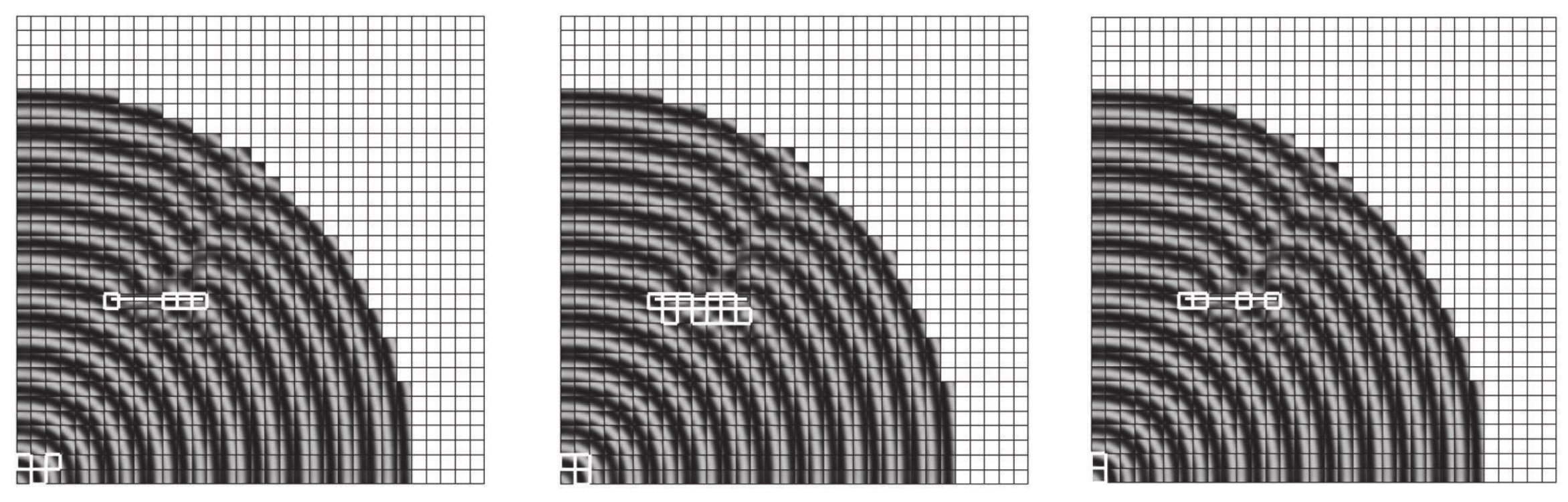}
	\caption{\emph{Effect of subsampling with random ($n_z \approx 20 \%$) patterns - Line defect.}}
	\label{Subsampling_Effects_32_32_Line}
\end{figure*}

Another \emph{visual} look to the variability introduced by random sampling is taken by applying sub-sampling to the line defect problem with a $32 \times 32$ region partition. Figure~\ref{Subsampling_Effects_32_32_Line} shows the saliency maps obtained using four randomly generated ($n_z \approx 20 \%$) sub-sampling patterns. While the results feature the expected variability, they consistently pinpoint the anomalous region, often capturing the length of the crack path as well as the thickness of the area of influence of the defect.

\renewcommand{\arraystretch}{1.5}
\begin{table}
\centering
\caption{\emph{Effect of sub-sampling on several metrics of saliency identification.  \jdh{There are $3$ true anomalies. Results presented here are averaged over $500$ random trials, each corresponding to a different random selection of sub-sampling locations, for several effective sub-sampling rates $n_z$.}}  }
\vspace{0.1in}
{\small \begin{tabular}{|c|c|c|c|c|c|}
  \hline
%
  $n_z$      & $50\%$ &  $33\%$ & $20\%$       & $10\%$                   & $7\%$         \\ \hline \hline
  Correct Discoveries           & $2.99$   & $2.96$ & $2.93$ &   $2.83$          & $2.37$  \\ \hline
  False Discoveries             & $1.27$   & $1.88$ & $2.46$ &   $4.32$          & $5.28$  \\ \hline \hline
  Correctly Disc. Regions          & $2.99$   & $2.96$ & $2.94$ &   $2.93$          & $2.72$  \\ \hline
  Falsely Disc. Regions             & $0.86$   & $1.14$ & $1.37$ &   $2.44$          & $3.27$  \\ \hline
   False Disc. $@$ Origin             & $0.83$   & $1.05$ & $1.24$ &   $2.14$          & $2.51$  \\ \hline
 \end{tabular}}
\label{Table_subsampling}
\end{table}

We also examine the performance of our procedure using deterministic sub-sampling sequences.  We select a family of sequences where the evaluation points sit on a double cross (horizontal-vertical and diagonal, corresponding to a sub-sampling ratio $n_z \approx 22\%$), where the topology choice is dictated by the attempt to have sufficiently dense sequences of sensing points along three important directions of propagation.  The results in Fig.~\ref{Full_Cross_Sub_sampling_3_Def_16_16} indicate that this sensing topology performs optimally in terms of correct saliency detection, as all the defects are identified without outliers. We can further increase the sub-sampling within the same family of pattern by further downsampling by a factor of two ($n_z \approx 11\%$) or four ($n_z \approx 6\%$) along each direction of the cross (Fig.~\ref{Full_Cross_Sub_sub_sampling_3_Def_16_16} and~\ref{Full_Cross_Sub_sub_sub_sampling_3_Def_16_16}). The results, as expected, decay, however the capability of the method to identify the correct salient regions is preserved, as a testament to the validity of the selected sequence pattern. It is worth noting that, although the cases of Fig.~\ref{Full_Cross_Sub_sub_sub_sampling_3_Def_16_16} and~\ref{Subsampling_3_Def_16_16_Rand_7} enforce roughly the same sub-sampling ratio, they profoundly differ in that one is random and the other one deterministic. As a result, although the outcomes plotted in Fig~\ref{Subsampling_Effects_16_16} happen to be comparable, the one of Fig.~\ref{Subsampling_3_Def_16_16_Rand_7} is subjected to the variability documented in Table~\ref{Table_subsampling}, which indicates an overall inferior performance. This result suggests that, while random sub-sampling represents an easy-to-implement and sufficiently reliable option, it may be possible to design deterministic sampling sequences with superior performance in the context of the proposed saliency-identification procedure.



\subsection{\jdh{Limits of sub-sampling for anomaly detection?}}

Our consideration of spatial subsampling approaches was motivated here by a desire to alleviate the computational burden on the data processing component of the inference procedure.  On the other hand it is, of course, well known that continuous-time bandlimited signals (such as the propagating wavefields here) can often be \emph{exactly} recovered from a collection of their discrete samples.  Indeed, suppose that the maximum (spatial) frequency present in the wave data is $f_{\rm max}$.  Then, spatial sampling at a rate of at least $2f_{\rm max}$ (the Shannon/Nyquist rate) is sufficient to enable \emph{exact} recovery of the full wavefield data.  For a multiband signal -- one whose frequency domain representation is comprised of a collection of possibly disjoint spectral bands -- the minimal sampling rate required for reconstruction corresponds to the total size (or Lebesgue measure) of its spectral support.  This condition, called the Landau sampling rate \jdh{\cite{Landau}}, is often a less restrictive condition than the Shannon/Nyquist rate.  \jdh{Indeed, several works have examined the recovery of multiband signals at sampling rates approaching the Landau rate -- see, for example, \cite{Cheung90, Cheung93, Vaidyanathan90, Feng96, Bresler96}.}

This suggests an alternative, \emph{brute force} approach, where the subsampled data is first used to \jdh{recover} the wavefield in the whole structure. Then, the full reconstructed wavefield is processed (e.g., using the proposed approach) to identify anomalous regions.  In light of this, performing an inference of structural anomalies from undersampled data is perhaps most interesting if the spatial sampling rate is sufficiently low, such that it would be impossible to reconstruct the wavefield from the sampled data.  How far are we from this ``sub-Landau'' sampling goal?

\begin{figure*}[t]
\centering
	   \subfigure[Defect-free plate]{\label{DFT_No_Defects} \includegraphics[scale=0.6]{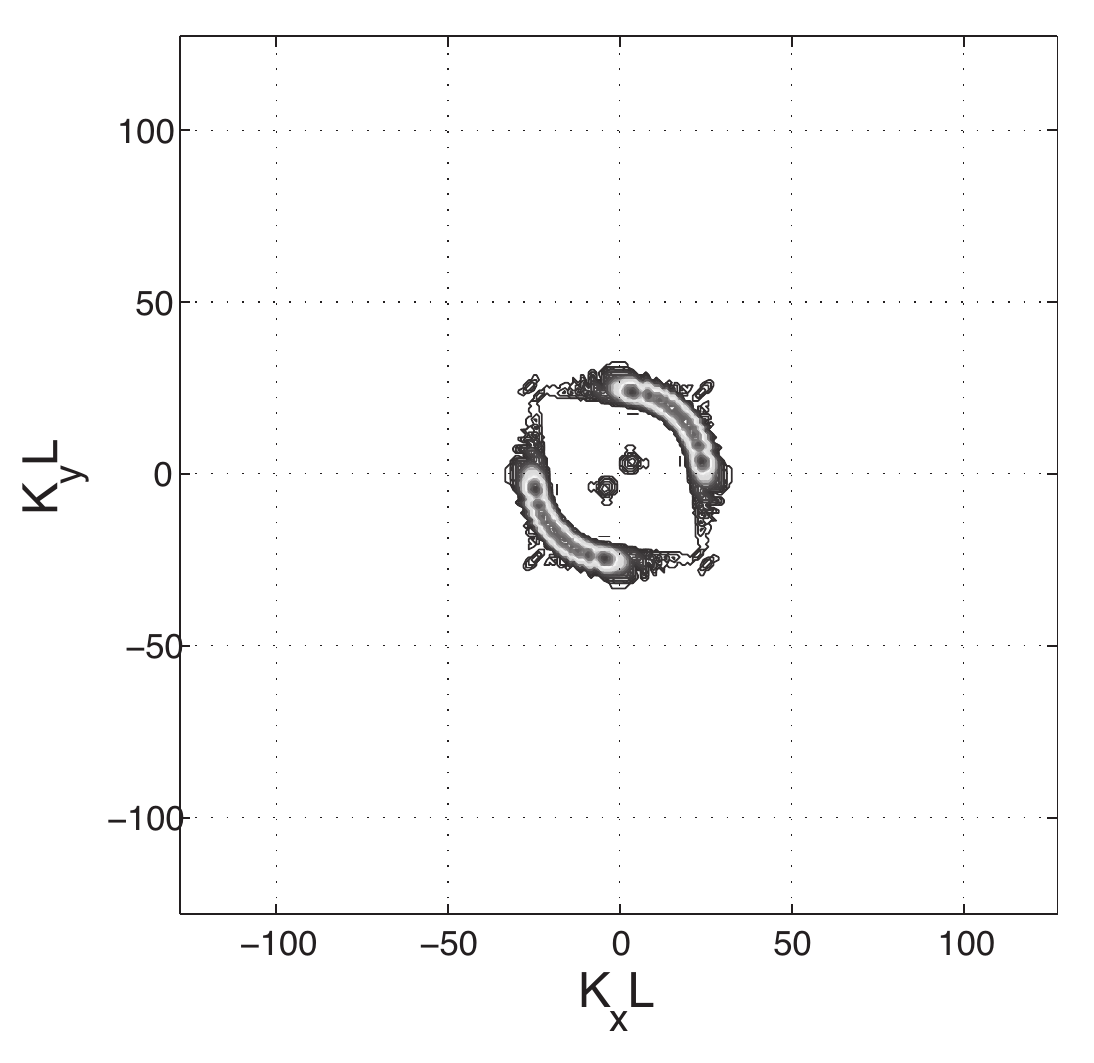}}
        \hspace{0.5in}
	   \subfigure[Scattered defects]{\label{DFT_Defects} \includegraphics[scale=0.6]{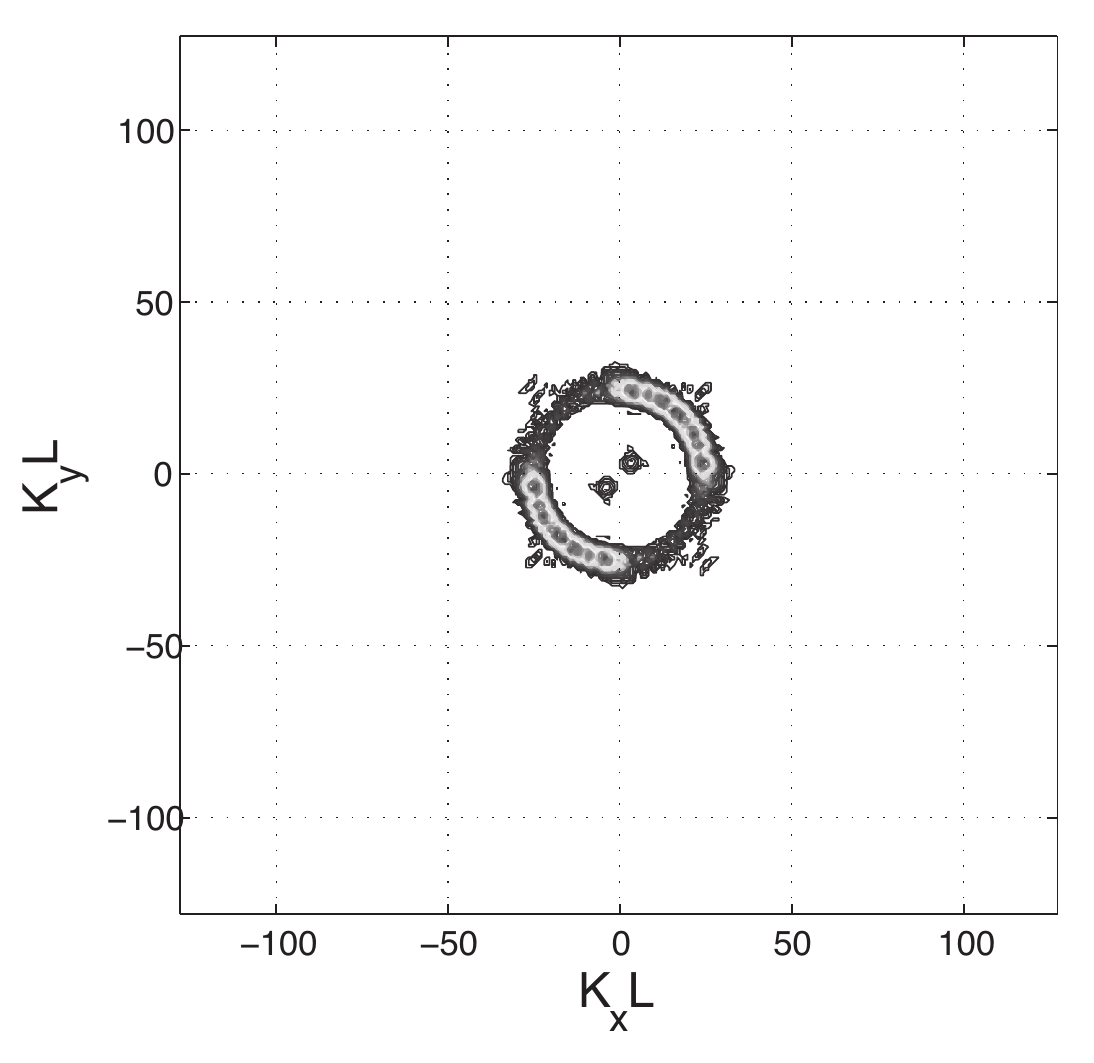}}
	\caption{\emph{$DFT$ charts revealing the Landau sampling rate limit and the signature of scattering-induced reflections.}}
	\label{DFT_Charts}
\end{figure*}

The Landau rate can be estimated by inspection of the signal in the spectral plane. Fig.~\ref{DFT_Charts} shows the spectral representation of the wave of Fig.~\ref{Def_3_Regions_8_8_Wavefield} with and without defects, obtained by taking the two-dimensional Discrete Fourier Transform (2D-DFT) of the wavefield $w(x,y,t_{f})$ and where $K_xL, K_yL$ are the components of the in-plane wavevector normalized by the size of the domain. Here, a back-of-the-envelope calculation of the area of the occupied spectrum, relative to the total area, shows that the Landau rate corresponds to downsampling the data to about 2.5\%. In contrast, here we saw reasonable performance only for downsampling rates at about 6-7\%, depending on the sampling strategy and the nature of the defect. This suggests that further improvements are necessary to achieve the goal of sub-Landau defect identification using our proposed approach.

We note that in real applications there may be advantages of adopting approaches like the one proposed here, which make inferences directly from the \jdh{sub-}sampled data.  \jdh{Namely}, as alluded above, by virtue of the reduced data set sizes, inferring anomalous regions from the \jdh{sub-}sampled data directly enjoys the benefit of lower computational complexity relative to the same approach applied to the full wavefield data. \jdh{Overall, it} remains to be seen whether accurate anomaly detection can be achieved by our method using sub-Landau sampling rates, \jdh{or whether the Landau rate represents a fundamental limit on the sub-sampling rate using our approach.} A full investigation \jdh{of these topics} is left for future work.

\section{Conclusions and recommendations for future work}\label{Conclusions and recommendations for future work}

This work investigates a saliency-based approach for the detection and localization of anomalies in two-dimensional structures probed with propagating waves. The concept of saliency is here associated with the notion of sparsity, and is interpreted in terms of deviations from a ``common'' linear model for local (in space and time) subsets of the propagating wavefield data. This concept is adapted to the wave problem and mathematically formalized as a PCA problem performed on partitions of the domain in conjunction with a time shift correction of the waves inside each region to remove ambiguity.  The detection capability of the method has been tested against two benchmark problems in damage detection (distributed defects and line defect) and is found to be a viable approach over a large range of region partitions and \jdh{spatial} sampling rates.  Indeed, the performance of the proposed method degrades gracefully as the sample rate decreases.

We note also that many recent efforts have proposed convex formulations of the low-rank plus outlier matrix decomposition problem \cite{Chandrasekaran11, CandesPCA, Outlier}, while our approach proposed here utilized simple matrix computations (low rank approximation by truncated SVD).  Other recent works have examined the low-rank plus outlier decomposition problem from limited or compressive measurements \cite{Waters11, WrightCPCP12} -- a task that is very similar in spirit to the subsampling approach presented in this work.  It will be interesting to see whether these convex formulations provide any improvements in our anomaly localization performance, and whether the theoretical foundations outlined in these works can provide additional insight into our problem. 

\jdh{It is worth noting that the computational aspect of our approach is reminiscent of the well-known DORT (D\'{e}composition de l`op\'{e}rateur de retournement temporel) detection approach \cite{Prada94, Prada95, Prada96}, in that the DORT approach also relies on eigenanalysis of certain matrices in order to locate defects.  There are, however, a few key differences between DORT and our proposed approach.

First, our proposed approach is based on detecting subtle differences or anomalies in the \emph{temporal} transient behavior of the propagating wavefield generated from a \emph{single} actuator, using multiple measurements obtained in the immediate vicinity of potential defects.  In contrast, the DORT approach utilizes multiple transmitters and receivers (transducers), and relies on a frequency domain analysis of the corresponding multiple input multiple output system.  In DORT, each path from transmitter to receiver is modeled as linear and time invariant (LTI) system, the essential idea being that the presence of pointlike scatterers in the medium will alter the responses of each of the channels in an identifiable way.  Our approach relies on the notion that defects or scatterers will locally (in space and time) perturb the wavefield as it interacts with the defect. (We note that frequency domain analysis of signals transmitted and received by multiple transducers also forms the basis of detection techniques based on the Multiple Signal Classification, or MUSIC, procedure -- see, for example, \cite{MUSIC12}.)

Further, and somewhat more subtly, the DORT approach implicitly assumes that the number of transducers be greater than the number of (resolvable) pointlike scatterers in the medium.  For instance, the DORT approach applied to a system with $L$ transducers prescribes describing the resulting multiple input multiple output system in the frequency domain (and at at a particular frequency) in terms of an $L\times L$ transfer matrix.  This matrix is learned or approximated via experimentation, and the number of nonzero eigenvalues of this matrix corresponds to the number of resolvable point-like scatterers.  Thus, DORT requires that the number of defects (say $d$) be smaller than $L$, else the results may be ambiguous.  Stated another way, if the rank of the transfer matrix in the DORT approach is less than the number of pointlike scatterers $d$ (as it would be in a single input multiple output system for any $d>1$), DORT will be unable to identify the $d$ distinct scatterers.  In contrast, we find in our initial investigations here that our approach succeeds in identifying the locations of multiple pointlike defects as well as crack-like defects using measurements of a propagating wavefield generated by only a single actuator. It is however important to emphasize again that, while it relies on a single actuation source, our method utilizes a significant number of sensing points to achieve the performance levels discussed above.}

Another interesting direction pertains to how the measured data is utilized in the inference procedure.  There is no inherent restriction to use the raw measured data itself in our inference approach; alternatively, we may consider some (perhaps nonlinear) \emph{preprocessing} of the measured wavefield data (e.g., conformal mapping, projective geometry, power flow statistics, etc.) in an effort to enhance anomalous features in the data.  Finally, it is intriguing to consider a \emph{multi-step} sampling and detection algorithm, where the proposed subsampling approach is used in several stages to iteratively identify smaller and smaller subsets of the material domain that may contain anomalies.  This sort of adaptive ``coarse-to-fine'' sampling strategy has been successfully employed in certain image processing tasks (e.g., \cite{Castro04}); it remains to be seen whether similar notions can be utilized here.  This is the objective of current investigations, an account of which is left for future work.

\section*{Acknowledgments}
J. Haupt acknowledges partial support by NSF Grant CCF-1217751.



\bibliographystyle{elsarticle-num-names}

\bibliography{NSF_bib}

\end{document}